\pdfoutput=1

\documentclass[11pt]{article}

\usepackage{acl}

\usepackage{times}
\usepackage{latexsym}

\usepackage[OT6,T1]{fontenc}    

\usepackage[utf8]{inputenc}

\usepackage{microtype}

\usepackage{inconsolata}

\usepackage{amsmath}
\DeclareMathOperator*{\argmax}{arg\,max}
\usepackage{booktabs}
\usepackage{tabularx}
\usepackage{bxcjkjatype}
\usepackage{graphicx}
\usepackage{caption}
\usepackage{subcaption}
\usepackage[russian,english]{babel}
\newcommand{\armenian}{\fontencoding{OT6}\fontfamily{cmr}\selectfont}
\DeclareTextFontCommand{\textarmenian}{\armenian}

\title{Where on Earth Do Users Say They Are?: \\Geo-Entity Linking for Noisy Multilingual User Input}

\author{Tessa Masis \normalfont{\emph{(they/them)}} \quad
    \bf{Brendan O'Connor} \\
    University of Massachusetts Amherst, MA, USA\\
    \texttt{\{tmasis, brenocon\}@cs.umass.edu}
  }

\begin{document}
\maketitle
\begin{abstract}
Geo-entity linking is the task of linking a location mention to the real-world geographic location. In this paper we explore the challenging task of geo-entity linking for noisy, multilingual social media data. There are few open-source multilingual geo-entity linking tools available and existing ones are often rule-based, which break easily in social media settings, or LLM-based, which are too expensive for large-scale datasets. We present a method which represents real-world locations as averaged embeddings from labeled user-input location names and allows for selective prediction via an interpretable confidence score. We show that our approach improves geo-entity linking on a global and multilingual social media dataset, and discuss progress and problems with evaluating at different geographic granularities. 
\end{abstract}

\section{Introduction}

The real-world geographic location of social media users is valuable data for many computational social science tasks, including disaster response \cite{kumar2019location}, disease surveillance \cite{lee2013real}, analyzing language variation \cite{huang2016understanding}, and comparing regional attitudes \cite{rosenbusch2020interregional}. Many studies have used Twitter (now known as X) data for such analyses, focusing on geo-tagged tweets where each tweet is associated with latitude and longitude coordinates. However, geo-tagging with coordinates was deprecated in June 2019 and even before then only a small percentage of tweets ($<2$\%) was geo-tagged \cite{kruspe2021changes}. 

It has thus become increasingly necessary to infer location from user profiles and especially from the free text Location field, in which a user may enter anything they want to identify their location. This field is frequently specified, with at least 40\% of users providing recognizable locations in over 60 different languages \cite{huang2019large}. 
The task of linking a location reference to the actual geographic location is known as \emph{geo-entity linking} (see Table \ref{tab:examples} for examples). There are few open-source multilingual geo-entity linking tools available and existing ones are often rule-based \cite{alex2016homing, dredze2013carmen}, which may break easily in noisy social media settings, or LLM-based \cite{zhang2023geo}, which are too expensive for large-scale datasets. 

\begin{table*}[t]
\centering
  \begin{tabular}{lll}
    \toprule
    \textbf{User-input location} & \textbf{Real-world location} & \textbf{Type of noise}\\
    \midrule
    TURKEY/SİNOP & Sinop, Sinop, TR & Uncommon punctuation use\\
    \begin{uCJK}\UTF{798F}\UTF{5CF6}\UTF{770C}\UTF{3044}\UTF{308F}\UTF{304D}\UTF{5E02}\end{uCJK} & Iwaki, Fukushima, JP & Non-Latin script\\
    Catskills & Hyde Park, New York, US & Informal/alternative name\\
    where the wild things are & N/A & Not a real location\\
    \bottomrule
  \end{tabular}
  \caption{Examples of user-input location references, the real-world locations they should be linked with, and the type of noise that the geo-entity linking model must be able to handle. }
  \label{tab:examples}
\end{table*}

In this paper, we investigate the task of geo-entity linking for noisy, multilingual user-input location references. Our work makes the following contributions:

\begin{itemize}
    \item We propose a method for geo-entity linking of noisy and multilingual user input by representing real-world locations with averaged embeddings from labeled user-input location names. Unlike previous methods, ours enables selective prediction via an adjustable threshold for cosine similarity scores, which we analogize with confidence scores (\hyperlink{section.4}{\S4}).
    \item We compare performance of multiple variations of our proposed method on a global and multilingual dataset, and show that all of them outperform the leading baseline (\hyperlink{section.5}{\S5}).
    \item Through a manual annotation experiment, we approximate accuracy upper bounds on our dataset and show that our method is near the upper bound at country- and administrative-levels but quite far below at the city-level. We discuss problems with geo-entity linking social media data at the city level (\hyperlink{section.6}{\S6}). 
\end{itemize}

\section{Related Work}

\textbf{Geo-entity linking}, also known as toponym resolution, seeks to link some mention of a geographic entity to the correct entity in a target database. Previous approaches typically use some combination of the mention's text and/or context; knowledge bases (e.g. gazetteer, Wikipedia) which contain features such as population, location type, etc.; and coordinates/geometric features. They may use some mix of rule-based, unsupervised, and supervised methods. The majority of prior work on geo-entity linking has focused on data in the English language and in the domain of news articles \cite{lieberman2012adaptive, speriosu-baldridge-2013-text, kamalloo2018coherent, cardoso2019using, kulkarni2021multi, cardoso2022novel, sa2022enhancing, li2022spabert, sharma2023spatially, zhang2023improving}. 
We note that the geospatially grounded model GeoLM -- which was trained only on English data -- was evaluated on geo-entity linking in a way similar to our proposed method, by ranking locations by cosine similarity between each candidate and the query location \cite{li2023geolm}. This is similar to our proposed NameGeo method, although the authors did not explore using cosine similarity thresholds for selective prediction or any of the other variations that we investigate.  

Some prior work has examined geo-entity linking in historical texts \cite{smith2001disambiguating, ardanuy2017toponym, ardanuy2020deep}, which includes English,  Spanish, Dutch, and German data; and in web pages \cite{moncla2014geocoding}, which includes French, Spanish, and Italian data. 

Most relevant to the current work are previous studies which have examined geo-entity linking in social media data. \citet{alex2016homing} uses a rule-based English system, \citet{dredze2013carmen} uses a multilingual rule-based system, and \citet{zhang2023geo} uses a multilingual LLM-based system. 

\textbf{Entity-linking} is the broader task of linking some mention of an entity -- which could be a person, place, or organization -- to the correct entity in a target database. We note a similarity between our proposed UserGeo method and one introduced for entity-linking in \citet{fitzgerald2021moleman}, in that they both represent entities using all mentions in the training data. However, our method is simpler as it does not involve model training. Additionally, UserGeo represents an entity by averaging mention embeddings instead of having a separate embedding for each mention, which induces a more holistic entity representation that can better handle noisy mentions present in social media data. 

\textbf{Predicting user location}, also known as user geolocation, is a task distinct from geo-entity linking social media data in that it seeks to determine the location of a user using both text data and user metadata, including post content, user bio, user language, time zone, or social networks \cite{han2012geolocation, jurgens2015geolocation, rahimi2015exploiting, huang2016understanding, rahimi2017continuous, izbicki2019geolocating, huang-carley-2019-hierarchical, luo2020overview}. The Location field is one of many features used for location prediction, if used at all.

\textbf{Geoparsing} is the task of both identifying and linking geographic entities in unstructured text \citep{wang2019enhancing}, essentially combining toponym recognition \cite{hu2023location} and geo-entity linking. Geoparsers focused on standard English texts (including news articles, Wikipedia, or scientific papers) include CLIFF-CLAVIN \cite{d2014cliff}, TopoCluster \cite{delozier2015gazetteer}, CamCoder \cite{gritta2018melbourne}, and DM\_NLP \cite{wang2019dm_nlp}. Other geoparsers include the Edinburgh geoparser for historical English text \cite{grover2010use}, GeoTxt for English social media data \cite{karimzadeh2013geotxt, karimzadeh2019geotxt}, and Perdido for French texts \cite{moncla2014geocoding}.

\section{Task and Data}

\subsection{Geo-entity linking task}

Given a target location database $D$, a training set $T$ containing user-input location name and ground truth location pairs, and a test set of user-input location names $I$, for each $i \in I$ we model

$$ \argmax_{d \in D} \text{score}( d, T, i) $$

to predict best matching geographic entity $d$, which is represented by a triple containing a city name, primary administrative region name (e.g. state, province), and country name. (The city and administrative region names may be empty strings if the entity is of a higher granularity, e.g. a country) The score() function evaluates the quality of $d$ as a match to user input $i$, given training data $T$. 

We note that a user-input location name $i$ may contain multiple locations or no real locations. Predicted entity triples are allowed to be composed of only empty strings (referred to as \textsc{Null}), indicating that no location could be predicted for the given user input.

\subsection{Data}

\textbf{Target location database. }We use a modified version of the GeoNames\footnote{\url{https://www.geonames.org/}} database, which contains entity names and coordinates for over 11M countries, administrative regions, counties, and cities across the globe. We filtered this database to exclude cities with populations under 15K, since tweets are more likely to come from more populated areas. Our final target location database contains 28,767 distinct locations: 252 countries, 3,947 administrative regions, and 24,568 cities. 

\textbf{Train and test dataset. }We use data from the \textsc{Twitter-Global} dataset \citep{zhang2022changes}. The original dataset is described as containing data from 15.3M tweets which are either tagged with geocoordinates or Twitter Place objects and are posted by users with a non-empty Location field. The tweets were posted from 2013 to 2021 and contain global and multilingual data. 

\begin{figure}[t]
    \centering
    \includegraphics[width=.48\textwidth]{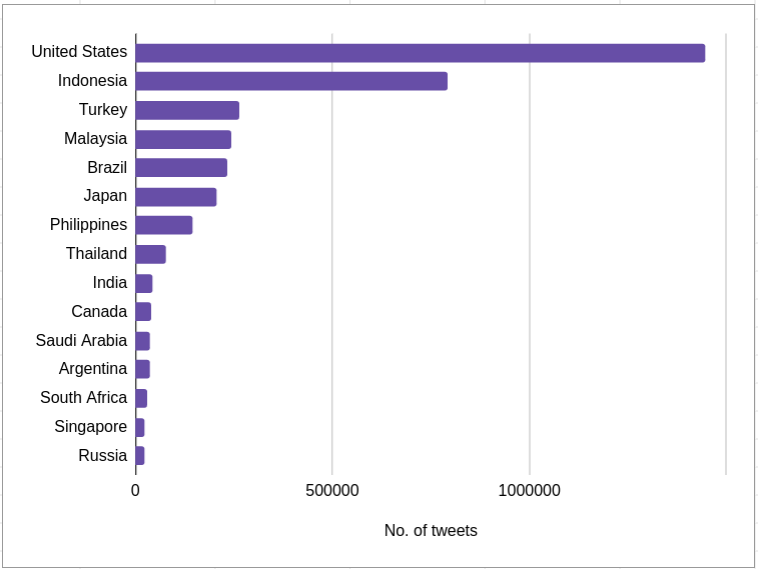}
    \caption{Tweet frequency for top 15 countries in the geocoordinate-tagged subset of \textsc{Twitter-Global}. } 
    \label{fig:country-distribution}
\end{figure}

We use only the 4.1M geocoordinate-tagged tweets in \textsc{Twitter-Global}, because geocoordinates are much more reliable than Place objects. Geocoordinates are meant to be the exact geolocation of the user's device and always specify a precise latitude and longitude, while Place objects are pre-defined geographic entities that the user selects from a list and may be at any granularity, from points of interest to countries \cite{kruspe2021changes}. Users may assign a Place to a post simply because they are talking about it and not because they are actually there. To identify the ground truth location for each geocoordinate-tagged tweet, we used the \emph{reverse-geocoder} library to map the coordinates to the closest city in our target location database.\footnote{\url{https://github.com/thampiman/reverse-geocoder}} There are tweets from 196 different countries, although they are not evenly distributed (see Fig. \ref{fig:country-distribution} for tweet distribution of the 15 most frequent countries). 

We note limitations to this data. Geo-tagged tweets are not a random or representative sample of all tweets and there have been shown to be biases in who uses geo-tagging. \citet{huang2019large} show that less than 3\% of Korean-speaking users geo-tag their posts while more than 40\% of Indonesian-speaking users do. Certain countries like Turkey and Indonesia have very high percentages of coordinate-tagged tweets out of all geo-tagged tweets (53\% and 67\%, respectively), where most countries have 15-30\%. In addition, users who have non-empty Location fields are more likely to use geo-tags, and there is strong homophily in geo-tagging behavior where users tend to connect to friends with similar geo-tagging preferences. \citet{pavalanathan2015confounds} show that users who geocoordinate-tag their posts versus have non-empty Location fields are measurably different groups in terms of demographics like age and gender. 

It has also been shown that a non-empty Location field does not always correspond to the geo-taggged coordinates. \citet{alex2016homing} find that in their dataset, 40\% of users have geo-tagged coordinates within 10km of their specific Location, 70\% are within 100km, and 85\% are within 1000km. Despite these limitations, previous work studying geo-entity linking and user geolocation has found geocoordinate-tagged data to be useful, so we use such data here.

\section{Methods}

\subsection{Proposed Method}

We propose a method (referred to as \underline{UserGeo}) that computes embeddings for each location in the target database and then, to predict a location for some user input, simply predicts the location with the closest embedding. To compute the target location embeddings, UserGeo uses training set $T$ which contains a pair $(x_t, y_t)$ for each tweet $t$, where $x$ is the user-input Location field and $y$ is the ground truth location triple. To help supplement locations with few user inputs, we additionally create a pair $(s(d), d)$ for each $d \in D$ where $s(d)$ is a string representation of $d$, with comma-separated city, primary administrative region, secondary administrative region, country name, and two-letter country code, as applicable. UserGeo creates embedding representation $Z_d$ for each location $d \in D$ by averaging all associated $x_t$ and $s(d)$ embeddings:

$$ Z_d = \frac{1}{|\{t:y=d\}|} \sum_{t: y = d} e(x_t) $$

where $e(x)$ is the embedding for $x$ from a pretrained language model. 

Then, for each user-input Location field $i \in I$, its predicted location triple $\hat{d}_i$ is the location embedding $Z_d$ that it has the highest cosine similarity with. If the cosine similarity with all location embeddings is below a given threshold $t$, then this is interpreted as a low confidence score and no prediction is made. In other words, 

$$ m = \max_{d \in D} c(Z_d, e(i)) $$
$$ \hat{d}_i = 
\begin{cases}
    \argmax_{d \in D} c(Z_d, e(i))& \text{if } m\geq t\\
    \text{\textsc{Null}} & \text{otherwise}
\end{cases} $$

where $c(a, b)$ is the cosine similarity between vectors $a$ and $b$. 

The motivation behind this method is that it leverages millions of examples of user-defined location names, essentially inducing a soft-alias location name database.\footnote{We don't investigate fine-tuned models, due to two main disadvantages. First, it is unclear exactly what the fine-tuned model would learn (e.g. learning regional slang, instead of variable ways of expressing a location). Second, it is more likely for the model to overfit to the training data and perform poorly on countries that are not well-represented in it. } We note parallels between our framework and two other methodological classes. First, Bi-encoders obtain separate embeddings for two sentences and then calculate cosine similarity between them. Second, Gaussian Discriminant Analysis classifies a point based on the minimum distance to clusters learned from training data. 

\subsection{Baselines}

We evaluate four variations of our method as baselines. \underline{NameGeo}: for each location $d \in D$, 

$$ Z_d = e(s(d)) $$

. In other words, a zero-shot version of UserGeo where each target location is represented only by its embedded string name.

\underline{Different embedding models}: we evaluate embedding with three \textsc{SBert} \cite{reimers2019sentence} variants -- the popular \emph{all-MiniLM-L6-v2} model, the multilingually-trained \emph{paraphrase-multilingual-miniLM-L12-v2} model, and the larger \emph{all-mpnet-base-v2} model -- and the geospatially grounded GeoLM \cite{li2023geolm}.

\underline{Variants}: there are multiple $s(d)$ functions to create different string representations of $d$ that are all included in $Z_d$. The original $s(d)$ represents $d$ as only "<city>, <admin2>, <admin1>, <country>, <2-letter country code>". Here, if $d$ is a country then it is also represented as "<country>"; if $d$ is a primary administrative region, then it is also represented as "<admin1>", "<admin1> in <country>", and as "<country> / <admin1>"; and if $d$ is a city, then it is also represented as "<city>", "<city> in <admin2> in <admin1> in <country>", "<admin1> / <city>", and as "<country> / <city>".  


\underline{Pruning}: removing outlier user inputs $x_t$ from each $Z_d$. A user input is determined to be an outlier if the squared Euclidean distance between $e(x_t)$ and $Z_d$ is farther than a given threshold.

We also evaluate the only prior open-source tool that was explicitly created for and evaluated on broadly multilingual data. \underline{Carmen 2.0}: uses a combination of regular expressions and manually curated aliases to predict real-world locations \citep{zhang2022changes}. 


\section{Experiments}

\subsection{Experimental setup}

\textbf{Data.} We divide the Twitter-Global data into a 90/10 split, with 3.7M examples in the training set and .4M in the test set. We evaluate at three levels of geographic granularity (city, primary administrative region, and country). A predicted triple is correct at the country level if the predicted country is a string match with the correct country; correct at the administrative level if both predicted country and administrative region are string matches with the correct ones; and correct at the city level if predicted country, administrative region, and city are all string matches with the correct ones. This hierarchical string matching procedure has limitations in that it does not remove all geographic ambiguity (e.g. if there are multiple cities with the same name in the same administrative region), but it should be effective in the vast majority of cases.

\begin{table}[t]
\centering
\begin{tabularx}{.48\textwidth}{l r r r}
Method & Country & Admin. & City \\
\hline\hline
\textsc{Carmen 2.0} & 43.5 & 27.3 & 9.8 \\
\hline
\emph{all-MiniLM-L6-v2}\\
\hline
NameGeo &  59.8 & 37.7 & 14.3 \\
UserGeo & \underline{\textbf{67.8}} & \underline{\textbf{44.2}} & \underline{14.8} \\
\hline
\emph{all-mpnet-base-v2}\\
\hline
NameGeo  & 60.9 & 38.3 & \underline{\textbf{14.9}}\\
UserGeo & \underline{67.4} & \underline{43.7} & 13.9\\
\hline 
\multicolumn{4}{l}{\emph{paraphrase-multilingual-MiniLM-L12-v2}}\\
\hline
NameGeo & 48.7 & 28.9 & 8.1\\
UserGeo & 57.0 & 34.3 & 9.4\\
\hline
\textsc{GeoLM}\\
\hline
NameGeo & 52.5 & 30.5 & 12.1 \\
UserGeo & 57.4 & 33.9 & 10.7 \\
\hline
\end{tabularx}
\caption{Accuracy scores (\%) at all three geographic granularities for Carmen 2.0, and NameGeo and UserGeo with different embedding models (all at threshold of 0, i.e. predictions are made for all examples). The highest score is bolded, the top 2 are underlined.}
\label{table:full-results}
\end{table}

\textbf{Metrics.} We use four metrics to evaluate methods. \emph{Accuracy} is the percentage of examples for which the method made a correct location prediction. Since many models do not always predict a location for all user inputs, we also evaluate \emph{coverage} (the percentage of examples with a non-\textsc{Null} prediction) and \emph{precision} (the percentage of correct non-\textsc{Null} predictions, following the standard definition of precision). 
For individual country performance we also evaluate F1-score, since most countries are uncommon. 

\begin{figure}[t]
  \centering
  \begin{subfigure}[ht]{.49\textwidth}
    \centering
    \includegraphics[width=\textwidth]{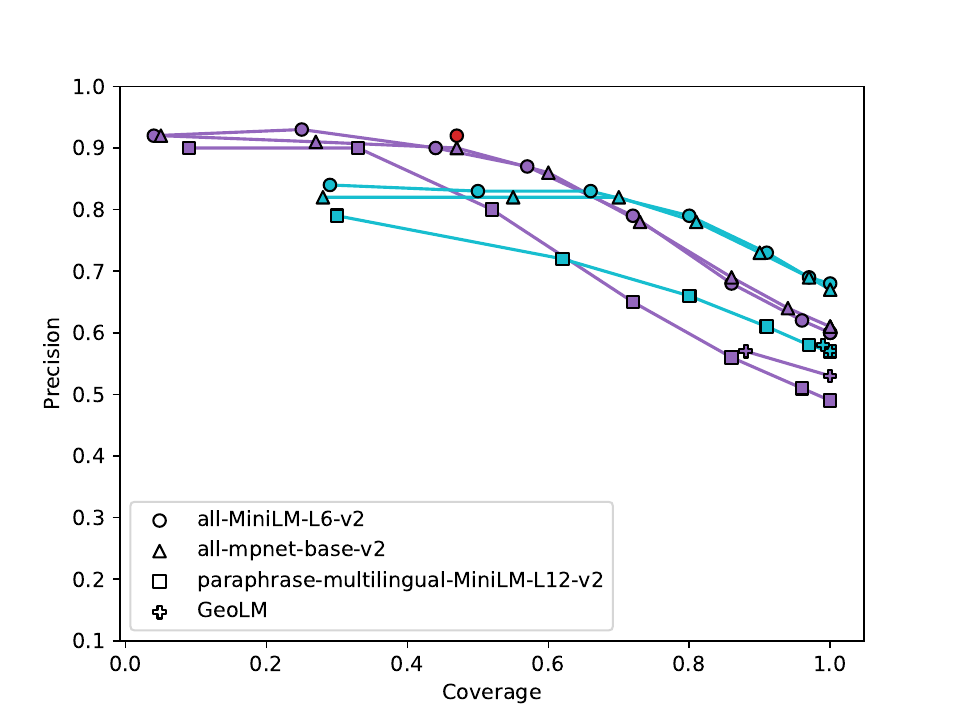}
    \caption{Country}
    \label{fig:country-curve}
  \end{subfigure}
  \begin{subfigure}[ht]{.49\textwidth}
    \centering
    \includegraphics[width=\textwidth]{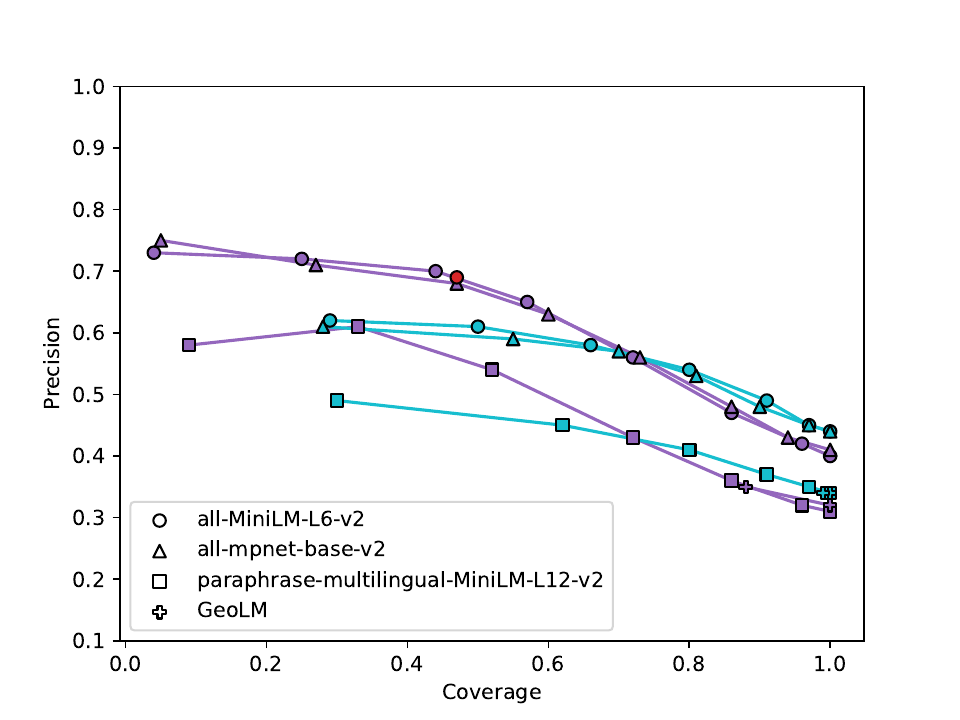}
    \caption{Admin.}
    \label{fig:admin-curve}
  \end{subfigure}
  \caption{Precision-coverage curves at the country (\textbf{a}) and administrative (\textbf{b}) levels. Red points are Carmen 2.0, purple are NameGeo, and cyan are UserGeo. NameGeo and UserGeo are shown with four different embedding models, where each point represents the precision and coverage at a threshold $t \in \{0, .1, .2, .3, .4, .5, .6, .7, .8, .9\}$. These curves demonstrate how thresholds can be used to tradeoff between precision and coverage with NameGeo and UserGeo. }
\label{fig:precision-coverage}
\end{figure}

\begin{table*}
\centering
  \begin{tabular}{lllll}
    \toprule
    \textbf{User-input} & \textbf{Carmen 2.0} & \textbf{NameGeo @0.5} & \textbf{UserGeo @0.5}\\
    \midrule
    TURKEY/SİNOP  & \textsc{Null} & "", Sinop, TR & Boyabat, Sinop, TR\\
    \begin{uCJK}\UTF{798F}\UTF{5CF6}\UTF{770C}\UTF{3044}\UTF{308F}\UTF{304D}\UTF{5E02}\end{uCJK} & \textsc{Null} & Zhongshu, Yunnan, CN & Iwaki, Fukushima, JP\\
    Catskills & \textsc{Null} & Catalca, Istanbul, TR & Greenburgh, New York, US\\
    where the wild things are & \textsc{Null} & \textsc{Null} & \textsc{Null}\\
    \bottomrule
  \end{tabular}
  \caption{Error analysis of the same user-input examples as in Table 1 (see Table 1 for corresponding real-world locations). Results from NameGeo and UserGeo are using the \emph{all-MiniLM-L6-v2} \textsc{SBert} model. Empty strings indicate that the model did not make a prediction at that geographic granularity, and \textsc{Null} indicates that no prediction was made. }
  \label{tab:error-analysis}
\end{table*}

\subsection{Results}

\textbf{NameGeo, UserGeo, and Carmen 2.0.} UserGeo achieves the highest accuracy at the country and administrative level, with gains over Carmen 2.0 of 25 and 17 points, respectively, and NameGeo achieves the highest accuracy at the city level, with gains over Carmen 2.0 of 5 points (Table \ref{table:full-results}). And while Carmen 2.0 has a competitive precision-coverage tradeoff, especially at the country and administrative level, its overall coverage is still quite low (see Fig. \ref{fig:precision-coverage}, and Fig. \ref{fig:prec-cov-city} in Appendix). In other words, it is often correct when it makes a prediction but it does not often make a prediction. It also only has a single precision-coverage point (similar to most other prior tools). In contrast, our proposed methods demonstrate the ability to tradeoff between precision or coverage by choosing a different threshold and can achieve a higher precision or coverage than Carmen 2.0 at certain thresholds. The threshold can be used to adjust the precision/coverage balance if, for a given application, it is more important to get predictions correct or if it is more important to make more predictions.\footnote{We note that these precision-coverage curves are valid for only this dataset. Users would likely have to reevaluate precision and coverage on a new dataset or domain in order to choose an appropriate threshold.}

We conducted a manual error analysis comparing Carmen 2.0, NameGeo, and UserGeo and observed trends in the types of errors made by each method (see Table \ref{tab:error-analysis}). First, Carmen 2.0 rarely makes predictions for user-inputs with unexpected punctuation or in non-Latin scripts. Second, NameGeo often incorrectly predicts locations that look superficially similar to the user input (e.g. it predicts a location in China for a user input written in Japanese, and a location named 'Catalca' for the user-input 'Catskills'). Third, UserGeo often correctly predicts locations for non-Latin inputs and alternate/informal location names. And lastly, all three models are frequently able to identify user-inputs that are not real locations. 

We compare country-level F1-scores\footnote{We use F1-score instead of accuracy here in order to better represent rare countries. } across countries for Carmen 2.0, NameGeo, and UserGeo, for the 23 countries with more than 1,000 examples in our test set, in order to investigate geographic bias in our models \cite{liu2022geoparsing}. We observe that the number of examples per country in the training set -- which may differ by multiple orders of magnitude -- does not appear to influence UserGeo performance (see Table \ref{table:f1-scores} in Appendix). This suggests that an unbalanced training set doesn't negatively impact performance as it might for a traditional supervised learning method.

\textbf{Different embedding models.} Across \textsc{SBert} bases, we find that the \emph{all-MiniLM-L6-v2} model surprisingly performs better than the multilingually-trained \emph{paraphrase-multilingual-MiniLM-L12-v2} model and it performs comparably with the larger \emph{all-mpnet-base-v2} model (Table \ref{table:full-results}, Fig. \ref{fig:precision-coverage}). Anecdotally, we found that the multilingual model performs worse for examples in Latin script; for example, NameGeo with the multilingually-trained model incorrectly predicts ("", Pukapuka, CK) for user input "Kucukyali izmir"  while NameGeo with \emph{all-MiniLM-L6-v2} correctly predicts (Izmir, Izmir, TR). On the other hand, it performs better for non-Latin scripts; UserGeo with the multilingually-trained model correctly predicts (Dnipro, Dnipropetrovsk Oblast, UA) for "\begin{otherlanguage*}{russian}Днепропетровск\end{otherlanguage*}" 
, while UserGeo with \emph{all-MiniLM-L6-v2} only partially correctly predicts (Pidhorodne, Dnipropetrovsk Oblast, UA) 
. We hypothesize that part of why the improved performance for non-Latin scripts does not outweigh the decrease in performance for Latin script is that non-Latin user inputs are a minority in this dataset; for a different dataset, it's possible that the performance of the multilingually-trained model may be better than that of \emph{all-MiniLM-L6-v2}. 

Regarding the GeoLM model, the only \textsc{SBert} model it outperforms is the multilingual one. Additionally, the cosine similarity threshold does not work effectively for GeoLM because the cosine similarities between a user input and each target location are very close; for example, the average cosine similarity between a correct prediction and an incorrect prediction for NameGeo with GeoLM at the country-level is .95 and .92, respectively. There is therefore a very limited range of values for which to have a threshold. Fig. \ref{fig:precision-coverage} demonstrates this limited range with very short precision-coverage curves for GeoLM.

Since the \emph{all-MiniLM-L6-v2} model performs the best out of all the embedding models, we use it for the rest of the experimental results.

\textbf{Variants.} Adding additional variants of the location name improves performance for NameGeo, and in fact NameGeo+variants does better than UserGeo at the city-level (see Table \ref{table:variation-results}). However, adding additional variants does not improve performance for UserGeo, and UserGeo+variants does marginally worse than UserGeo.

\begin{figure}[t]
  \centering
  \includegraphics[width=.49\textwidth]{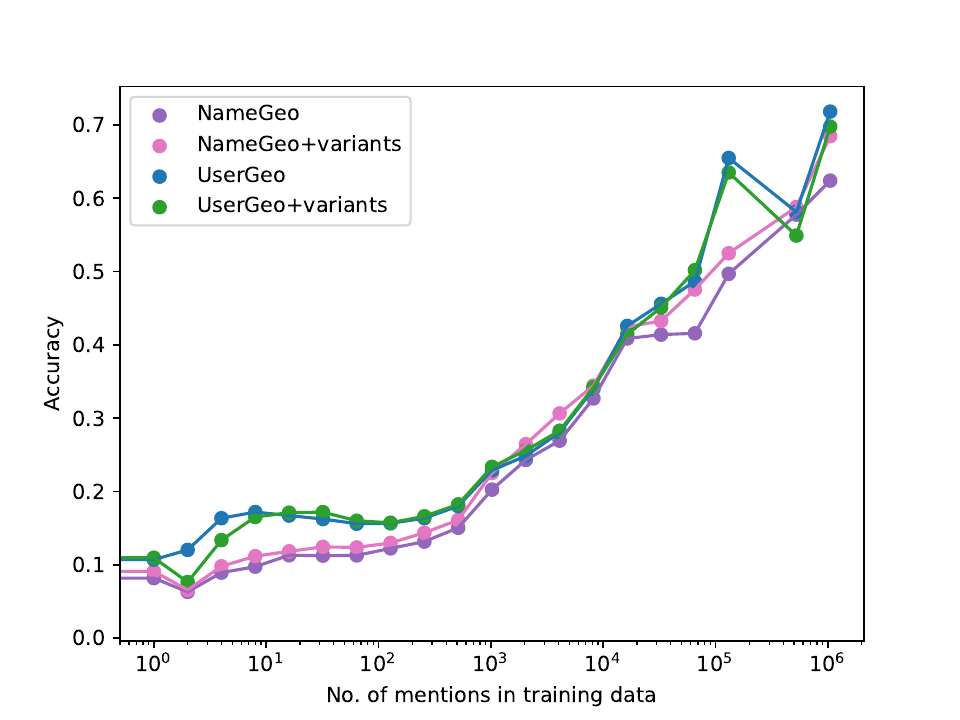}
  \caption{Average accuracy for a given number of mentions in training data (bucketed by $\left \lfloor{\log_2 \text{\#mentions}}\right \rfloor$),
  for NameGeo and UserGeo with and without adding location name variants. Location entities at all 3 geographic granularities are present in the plot. }
\label{fig:usergeo-mentions}
\end{figure}

We investigate the impact of number of location mentions in the training data on model performance. Fig. \ref{fig:usergeo-mentions} contains a plot of the number of mentions in the training data per location versus NameGeo/UserGeo accuracy. NameGeo does not use any user inputs from the training data so it can be considered a control for how training data affects performance in UserGeo. We see that while some training data is better than none (UserGeo consistently outperforms NameGeo when number of mentions is less than 1,000), it is not true that more training data always continues to improve performance (UserGeo and NameGeo have comparable performance when number of mentions is greater than 1,000). 
This suggests that having more variants or examples in a location's averaged embedding is helpful, but only to an extent. Few-shot is better than zero-shot, but after a certain point there are diminishing returns.


\textbf{Pruning.} We defined an outlier as an embedding that was more than the average distance from its centroid, where the average is calculated using all embeddings associated with the given location. On average, 38\% and 42\% of embeddings in each cluster were pruned for UserGeo and UserGeo+variants, respectively.\footnote{We determined through preliminary analyses that the average distance was an appropriate threshold, e.g. with a threshold of 2x the average distance, only 3\% of embeddings in each cluster were pruned on average and there was no improvement in performance. } 

In general, removing outliers did not improve performance and frequently made it worse (Table \ref{table:variation-results}). This suggests that having more variety in the averaged embeddings is good for performance, even if it comes at the cost of noise. 

We observed via error analysis that it is often the case that user inputs close to the centroid will be standard English spellings of the location, ones within 1x average distance will include the location spelled in different scripts or nearby/related locations, and ones more than 1x average distance will not be very related to the location. For example, for the country Armenia, user inputs less than .5x average distance include "Yerevan, Armenia" and "ARMENIA"; user inputs less than 1x average distance include "Armenia | \textarmenian{Հայաստան} | 
\begin{otherlanguage*}{russian}Армения\end{otherlanguage*}
" and "Azerbaijan, Baku"; and user inputs more than 1x average distance include "Paris/Singapore" and "Worldwide".

\begin{table}[t]
\centering
\begin{tabularx}{.48\textwidth}{l r r r}
Method & Country & Admin. & City \\
\hline
NameGeo &  59.8 & 37.7 & 14.3 \\
+variants & 62.0 & 40.9 & \underline{\textbf{17.0}} \\
UserGeo & \underline{\textbf{67.8}} & \underline{\textbf{44.2}} & 14.8 \\
+pruning & 63.5 & 41.4 & 13.2\\
+variants & \underline{66.0} & \underline{43.7} & \underline{15.3} \\
+variants+pruning & 65.2 & 43.4 & 13.9\\
\hline
\end{tabularx}
\caption{Accuracy scores (\%) for variations of NameGeo and UserGeo (all at threshold of 0); the highest score is bolded, the top 2 are underlined. The \emph{all-MiniLM-L6-v2} \textsc{SBert} model was used for all results. }
\label{table:variation-results}
\end{table}

However, it is also frequently the case that user inputs one may want to exclude are less than 1x average distance and inputs one may want to include are more than 1x average distance. For example, "Moscow" is less than 1x average distance and "ARMENI, ABOBYAN." is more than 1x average distance. User inputs that are semantically relevant (i.e. refer to the location) but stylistically dissimilar (e.g. contain emojis or uncommon punctuation uses) are often farther from the centroid than user inputs that are less semantically relevant but more stylistically similar (e.g. "New York" is within 1x average distance of the centroid for the country Gibraltar, while "G I B R A L T A R" is not). These results suggest that the model is not able to effectively differentiate between good noise and bad noise, and thus it is better to not do pruning and to keep as many user inputs in the training data as possible.

\section{Accuracy Upper Bounds}

To approximate an upper bound for accuracy scores on our dataset, we conducted an analysis of a random sample of user inputs to determine how many contained references to actual locations. The percentage of examples which contain a location at each granularity suggest an upper bound for accuracy, indicating the approximate proportion that a geo-entity linker could be expected to connect to a real geographic location. The first author manually annotated 120 random examples from the training set for whether or not the user input contained an identifiable reference to a location at the country, administrative, and/or city levels.\footnote{We did not count coordinates as containing a location, as we wanted to investigate the ability of the model to identify natural language location references. If a location was named that could be either an administrative region or a city, then the example was marked as containing both.}  

We include results from this annotation in Table \ref{table:manual-annotation} as well as a comparison with our best performing models' accuracy scores. UserGeo is only 5 points below the country-level upper bound and 14 points below the administrative-level upper bound, indicating that our current performance is fairly close to the upper limit in terms of accuracy. In contrast, NameGeo+variants is more than 30 points below the city-level upper bound.

We discuss a few problems with evaluating geo-entity linking at the city level for social media data. First, the assumption of the geocoordinates as ground truth is frequently untrue and especially so at the fine-grained city level. Unlike news articles or Wikipedia data, the place that a user puts on social media is not necessarily the place where they actually are. As \citet{alex2016homing} have shown, only 70\% of users have geocoordinates within 100km of the place specified in their Location field. Second, since we define ground truth as the closest city to the geocoordinates with a population higher than 15K, there will always be a mismatch between ground truth and Location field if a user puts a city with a lower population. Third, there is a disconnect between what is considered a city by a database such as GeoNames versus by an everyday person. Istanbul, Jakarta, Moscow, Gaborone, Lampang, and Santa Ana are all names of both cities and administrative regions in GeoNames -- when users put them in their Location field, it's unclear which one they are referring to. 

We hypothesize that these problems contribute to the poor performance of geo-entity linking tools at the city level. We additionally acknowledge how inferring location at the city level is more invasive than inferring at the country or administrative level because it predicts more fine-grained information about the user. Therefore, due to mismatches between geocoordinates and user-defined Location fields as well as the privacy concerns, we suggest that researchers only predict city-level location if it's necessary for a given application and otherwise use country- or adminstrative-level predictions.

\begin{table}[t]
\centering
\begin{tabularx}{.48\textwidth}{l r r r}
 & Country & Admin. & City \\ 
\hline
Upper bound & 72.5 & 58.3 & 49.2\\
\hline
NameGeo+variants & 62.0 & 40.9 & \textbf{17.0} \\
UserGeo & \textbf{67.8} & \textbf{44.2} & 14.8 \\
\hline
\end{tabularx}
\caption{Accuracy upper bounds and scores (\%) for our two best models, with highest scores bolded.}
\label{table:manual-annotation}
\end{table}

\section{Conclusion and Future Work}

In this paper we introduced new methods for geo-entity linking noisy, multilingual social media data with selective prediction. Of our two best performing methods, one does not require any training data (NameGeo+variants) while the other achieves state-of-the-art performance at the country and administrative levels (UserGeo). We also discussed problems with geo-entity linking at the city level for social media data, and suggested against doing this unless necessary for the application.

This work is not without limitations. We do not compare performance with LLM prompting methods, as they are too expensive for large-scale datasets and it's currently unclear how much prompt testing is necessary for a robust evaluation. However, it would be a useful comparison to know how LLMs perform at zero- or few-shot geo-entity linking; we leave this for future work. We also acknowledge that our methods rely heavily on \textsc{SBert} models, and thus are reliant on a third party for sustaining them. 

We plan to release a version of the UserGeo location embeddings in future, although we will not release the current version due to concerns about the \textsc{Twitter-Global} data (e.g. unbalanced across countries, unclear sampling methodology). We also hope to further evaluate our method on other domains with noisy location references, such as historical data, and to explore extensions of our method 
so it may be used for the broader task of geoparsing unstructured text.

\section{Ethical Considerations}

We discuss here ethical issues that may arise from 
using any geo-entity linking tool on social media data for a downstream application. When using a geo-entity linking tool on social media data, there is always the risk of de-anonymizing users through the inference of sensitive location information. It is generally recommended to use the lowest granularity necessary for the application simply because it is easier to correctly obtain than more fine-grained data \cite{kruspe2021changes}, but it is also true that lower geographic granularities protect individuals more due to being part of a larger aggregate \cite{dupre2022geospatial}. We also acknowledge the increasing importance of geomasking techniques, which aim to protect the privacy of individuals while preserving spatial information in geodata. \citet{lorestani2024privacy} survey the privacy risks of geocoded data and present a taxonomy of current geomasking techniques, and \citet{gao2019exploring} specifically examine the efficacy of geomasking techniques for protecting the privacy of Twitter users.

\section*{Acknowledgements}
We would like to thank the anonymous reviewers for their helpful comments and feedback. 
This work was supported by a National Science Foundation Graduate Research Fellowship (No. 1938059).

\bibliography{custom}

\newpage
\appendix

\section{Appendix}
\label{sec:appendix}

\begin{figure}[ht]
  \centering
  \includegraphics[width=.49\textwidth]{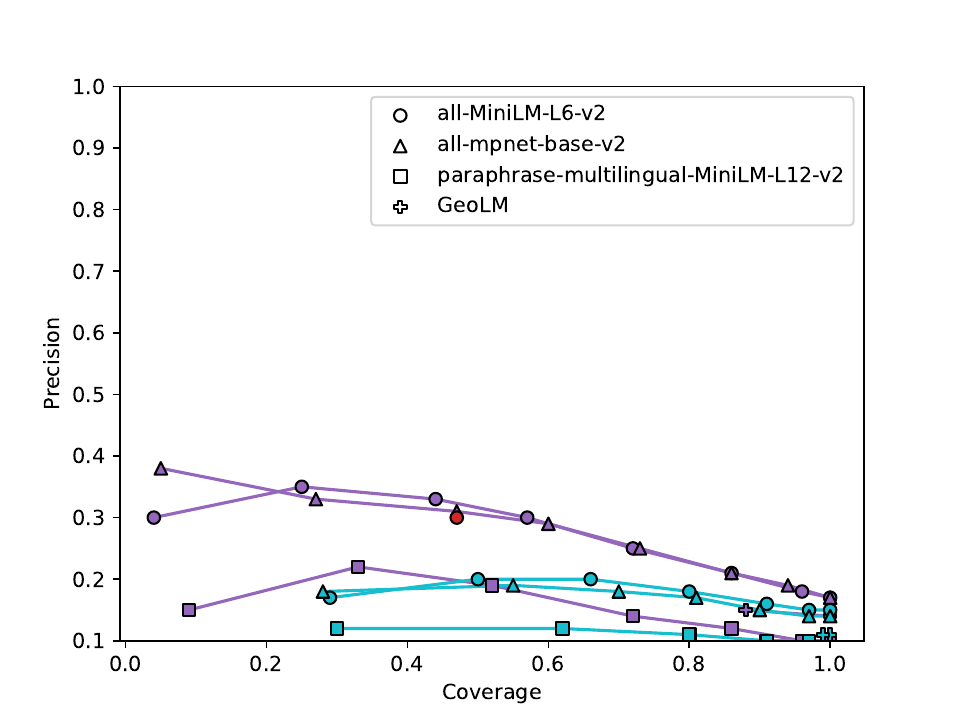}
  \caption{Precision-coverage curve at the city level. The red point is Carmen 2.0, purple points are NameGeo, and cyan are UserGeo. }
\label{fig:prec-cov-city}
\end{figure}

\begin{table*}[ht]
\centering
\begin{tabularx}{.8\textwidth}{l r r r r}
Country & No. mentions in train data & Carmen 2.0 & NameGeo & UserGeo \\
\hline
United States & 1300869 & .73 & .77 & .84 \\
Indonesia & 713967 &.54 & .73 & .74 \\
Turkey & 237472 & .57 & .91 & .91\\
Malaysia & 218257 & .49 & .70 & .70\\
Brazil & 210428 & .47 & .78 & .89\\
Japan & 184647 & .27 & .48 & . 92\\
Philippines & 131152 & .57 & .69 & .73\\
Thailand & 70403 & .55 & .67 & .64\\
Singapore & 43996 & .51 & .54 & .53\\
India & 40656 & .80 & .91 & .93\\
Canada & 36990 & .54 & .83 & .78\\
Saudi Arabia & 34559 & .40 & .48 & .73\\
Argentina & 34104 & .54 & .72 & .84\\
South Africa & 27228 & .59 & .77 & .80\\
Russia & 21820 & .53 & .92 & .91\\
Kuwait & 17829 & .48 & .65 & .59\\
Australia & 14363 & .53 & .85 & .86\\
Chile & 14268 & .80 & .87 & .89\\
Nigeria & 12482 & .80 & .84 & .87\\
Spain & 11699 & .40 & .65 & .79\\
Egypt & 10979 & .57 & .72 & .78\\
UAE & 10918 & .43 & .71 & .67\\
Pakistan & 9423 & .79 & .86 & .89\\
\hline
\end{tabularx}
\caption{Per-country country-level F1-scores for the 23 countries with over 1,000 examples in the test set. Results from NameGeo and UserGeo are using the \emph{all-MiniLM-L6-v2} \textsc{SBert} model. Both NameGeo and UserGeo outperform Carmen 2.0 for each of the 23 countries. We also note that F1-score can vary significantly by country, demonstrating the importance of evaluating per-country performance when using a geo-entity linker to investigate social media users who are from or identify with particular countries.}
\label{table:f1-scores}
\end{table*}

\end{document}